SAFE CONTROL TRANSITIONS: MACHINE VISION BASED OBSERVABLE READINESS INDEX AND
DATA-DRIVEN TAKEOVER TIME PREDICTION

**Ross Greer**
**Nachiket Deo**
**Akshay Rangesh**
**Mohan Trivedi**
Laboratory for Intelligent & Safe Automobiles[1]
University of California San Diego
USA

**Pujitha Gunaratne**
Toyota Collaborative Safety Research Center
USA

Paper Number 23-0331

## ABSTRACT

To make safe transitions from autonomous to manual control, a vehicle must have a representation of the awareness of driver state; two metrics which quantify this state are the Observable Readiness Index and Takeover Time. In this work, we show that machine learning models which predict these two metrics are robust to multiple camera views, expanding from the limited view angles in prior research. Importantly, these models take as input feature vectors corresponding to hand location and activity as well as gaze location, and we explore the tradeoffs of different views in generating these feature vectors. Further, we introduce two metrics to evaluate the quality of control transitions following the takeover event (the maximal lateral deviation and velocity deviation) and compute correlations of these post-takeover metrics to the pre-takeover predictive metrics.

## INTRODUCTION

It is important to plan for safe operation of intelligent vehicles in situations of system failure. Intelligent and autonomous vehicles face challenges when dealing with long-tail events, defined as events which occur with little to no regularity and are thus difficult for the dominant regime of learning-based perception and control models to operate safely. When such situations are identified, the vehicle may benefit from passing control to the human driver. However, there is risk in such control transitions, especially if the driver is not alert to the scene or ready to operate the vehicle; in these cases, it may be safer for the vehicle to perform an emergency maneuver such as braking or pulling over.

To mediate between these options, predicting the driver's takeover readiness is a critical human factor consideration for safe control transitions in conditionally autonomous vehicles. The duration of such a transition is quantified by the so-called Takeover Time (TOT), which measures the interval between an autonomous vehicle's request for manual driving (Takeover Request or TOR) and the time the driver assumes control. An illustration of the role of Takeover Time when an autonomous vehicle encounters an on-road hazard (and the necessary computation of the driver's ability to safely perform that takeover) are provided in Figure 1.

---

[1] cvrr.ucsd.edu

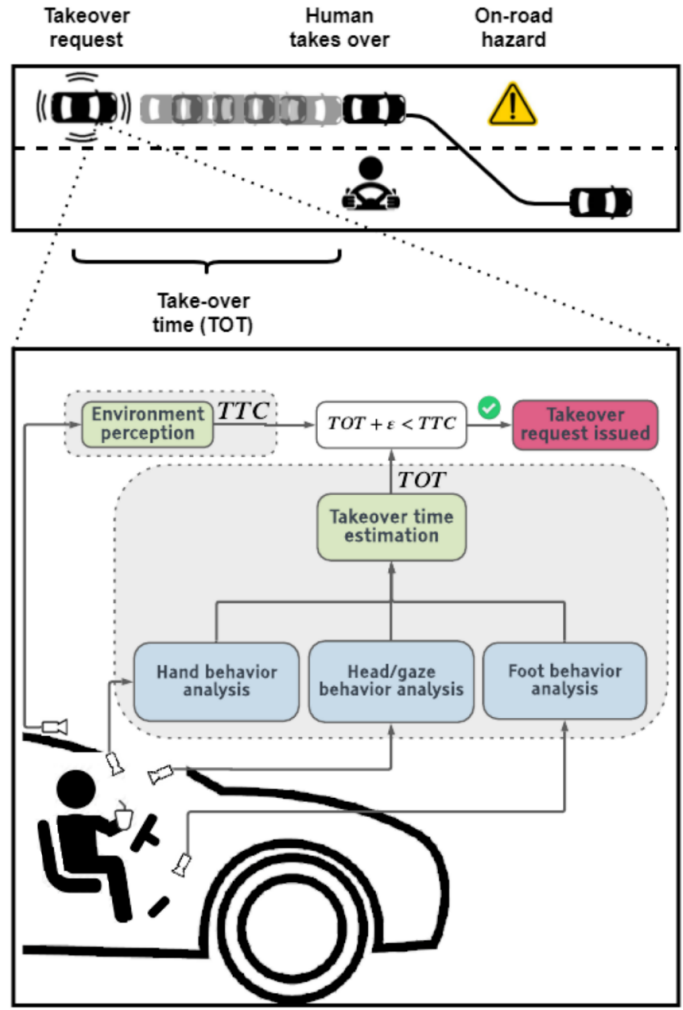

***Figure 1. An autonomous vehicle may issue a takeover request when encountering an on-road hazard. For a safe control transition, the vehicle must have an understanding of the driver's state, which can be inferred via in-cabin cameras observing the driver for visual cues.***

Establishing metrics to evaluate the effectiveness of a control transition is critical for effective experimentation and design of the systems we trust to make these decisions safely. Ego-vehicle metrics based on lateral and longitudinal motion immediately post-takeover provide information on how smoothly and safely the driver has established manual control.

Altogether, the problem of predicting readiness for control transition requires metrics across a before-during-after framework: how ready is the driver to take control (before), how long will it take the driver to establish control (during), and how successfully did the driver take control (after).

**RELATED RESEARCH**

Driving in real freeway and urban environments can subject drivers (and autonomous agents) to a variety of complex scenes, such as unexpected surround vehicle path changes and marked and unmarked intersections. Many research works seek to better analyze and understand the surrounding scene to make for safer autonomous interactions [1, 2, 3, 4], but of equal importance is the ability of the vehicle to provide transition of control to an alert driver to complete interactions under uncertainty. Driver state monitoring [5, 6] has many safety applications in crash prevention and mitigation under manual control, but here we focus on situations where vehicles begin in an autonomous state and return control to a human driver for safety. Commercial trends show growth in the use of driver-facing camera systems to facilitate this state monitoring, as both safety benefits of such systems and privacy-preserving mechanisms [7] improve. However, a recent study [8] determined that cars with such systems exhibited both inconsistent and unsafe behaviors as well as poor driver alerting on road departure and construction zone test contexts for a highly-automated vehicle. Further research shows that even when a takeover is enacted, there is significant risk of accidents even after driver intervention [9], though research has guided effective HMI to communicate the vehicle state (and request for manual control) to the driver [10, 11, 12]. Efforts towards improving safety in takeover situations align with the Fallback (Minimal Risk Condition) category in the NHTSA framework for Automated Driving Systems [13], further exemplified in the ADS test framework proposed by Thorn et al. [14].

Deo and Trivedi [15] define the Objective Readiness Index (ORI), a metric which quantifies a driver's readiness behind the wheel by normalizing and averaging ratings assigned by multiple human observers viewing feeds from in-cabin cameras capturing the driver's gaze, hand, and foot activity. Research connecting these cues to driver attention have provided an important basis for driver state analysis in safety applications [16, 17]. They show this metric can be predicted using an LSTM-based machine learning model inferring on observations from similar in-cabin camera feeds. Recent successful predictive methods encode such in-cabin camera observations into class probability vectors using convolutional neural networks. These include Rangesh and Trivedi's *HandyNet* CNN [18], Yuen and Trivedi's part affinity fields approach [19], Vora et al.'s gaze CNN [20], and Rangesh and Trivedi's forced spatial attention approach [21]. Such vectors serve as low-dimensional representations of the predictive information from appearance-based models of hand, eye, and foot activity of the driver.

Extending to an additional objective metric, Rangesh et al. [22, 23] show that computer vision and machine learning algorithms can be used to predict quantities such as the takeover time for a driver during a transition from autonomous to manual control. Both ORI and TOT estimation methods use as features the estimated positions of the driver's eye gaze, hands, and feet relative to the driving scene, steering wheel, and pedals respectively.

While many studies analyze driver takeovers from simulation, our research is conducted over naturalistic driving data collected from real autonomous vehicles operating on an experimental test track. Shi and Bengler [24] provide analysis of takeover times in relationship to external conditions and in-cabin driver tasks; we provide similar analysis and include learning-based methods of estimating a driver's readiness and takeover time.

Previous works demonstrate the ability of such algorithms to operate within a fixed experimental camera view; here we explore whether algorithms which predict takeover readiness generalize well to multiple driver views. Further, we explore the relationship between takeover readiness and the timing and *quality* of such takeover events. Takeover quality is of critical importance, as there is apparent difference in the reflexive establishment of motor readiness versus cognitive processing of a road situation which can be impaired by driver distraction [25].

**METHODS**

While related works collect gaze and hand features from a single camera view, here we adopt a multi-camera framework for evaluation. We apply convolutional models to estimate driver gaze and hand states from a variety of camera views, then further illustrate the effectiveness of these model outputs in estimating ORI and TOT, following the machine learning framework defined in [15, 22, 23].

**Models For Driver Behavior Analysis**

To analyze driver behavior in estimating ORI, we first collect frame-wise output features of driver gaze and hand activity, using the pipeline depicted in Figure 2. The method we employ begins with detection of the driver in each of the four camera views, followed by application of a human-pose estimation model to predict the driver's joint locations. Such an approach is traditionally referred to as "Top-Down". It is worth noting that these joint detection models are sensitive to the accuracy of the initial detection of the driver, and likewise the models for driver activity classes are sensitive to the accuracy of the detected joints.

For driver detection, we employ the MMDetection [26] implementation of Faster-RCNN [27] with Feature Pyramid Networks [28], using a ResNet-50 backbone [29] pretrained on COCO [30]. For joint detection, we employ the MMPose [31] implementation of HRNet [32], also pretrained on COCO human detections fit to a resolution of 256x192. We use the driver's ears, eyes and nose key-points to localize and crop their eye region. The cropped image is then passed through a convolutional neural network that classifies the driver's gaze zone. Similarly, we use the driver's elbow and wrist key-points to localize their hands. Cropped images of the driver's hands are passed through convolutional neural networks that classify the driver's hand location (e.g. on wheel, interacting with infotainment, in-air gesturing, etc.) and held objects (e.g. phone, beverage, etc.). Figure 3 shows the driver's eye and hands localized using their pose keypoints.

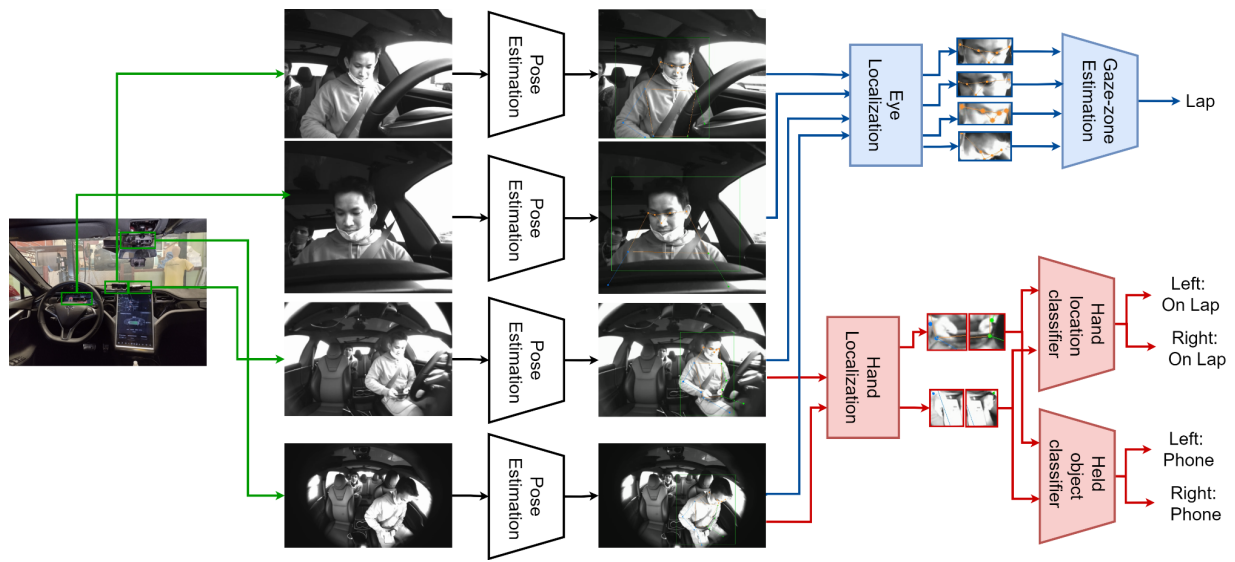

***Figure 2. Overview of driver behavior analysis: We apply a detector and human-pose estimation model to each of the four views to localize the driver and their joints. We use the pose key-points to localize the driver's eyes and hands. Cropped images around the driver's eyes and hands are passed through convolutional neural networks to classify the driver's gaze zone, hand location, and held object.***

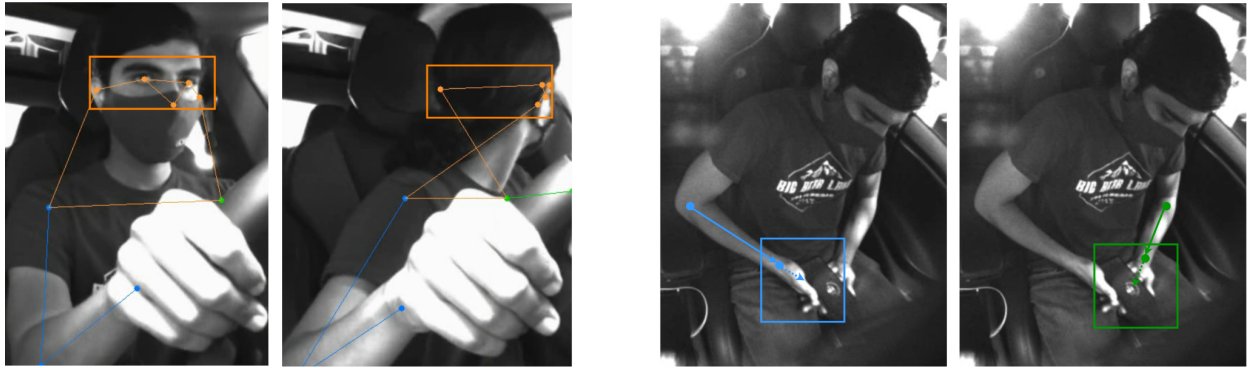

*Figure 3. Eye and hand localization: We localize the drivers eyes and hands using the estimated head, elbow, and wrist key-points.*

**<u>Dataset</u>** We collected gaze data from 9 different subjects for training the gaze and hand zone estimation models. Each subject was instructed to look at the different gaze zones sequentially by an experimenter. The subjects were encouraged to vary their head and body pose while ensuring that their gaze was directed towards the same gaze zone. Likewise, subjects were asked to place their hands through five hand locations: on steering wheel, on lap, in air (including gesturing), using infotainment unit, and on cupholder, followed by three held object activities: phone, tablet, and beverage (water bottle). The captured videos were then split into contiguous segments for each gaze zone, hand zone, and held object, providing labeled training data for the models. All 4 cameras synchronously captured the training data, yielding a total of 86,953 frames with labeled gaze zones, 181,584 frames with labeled hand zones, and 263,166 images per camera with labeled held objects. During training for the gaze and hand classification systems, we employed a cross-validation method whereby one subject is entirely removed from the training data, and used only as the evaluation subject. This technique helps in identifying the generalization of the models to subjects unseen in training data (i.e. avoiding overfitting to particular participants' hands and eyes).

**<u>Gaze Zone Classification</u>** We use the cropped image around the driver's eyes to classify the driver's gaze direction into one of five different gaze zones. We consider the following gaze zones related to driving as well as non-driving activities: forward, rearview (generalized to include left shoulder/blind spot, left mirror, rearview mirror, right mirror, and right shoulder/blind spot), lap, speedometer, and infotainment. Figure 4 shows an illustration of the driver looking at each of the 5 gaze zones in all 4 views. These set of gaze zones capture non-driving related tasks (NDRTs) such as interacting with the infotainment unit or using a handheld device, as well as an attentive driver checking the vehicle's surroundings in the front or rear.

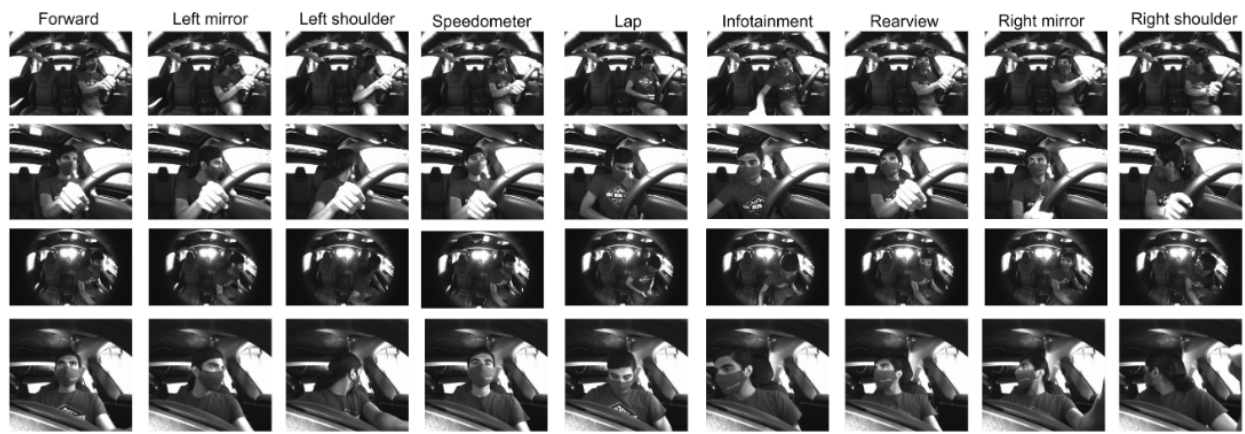

*Figure 4.* **The 5 gaze zones as seen from our 4 camera views.**

We used an EfficientNet-B3 [33] model for classifying gaze zones. We used the ImageNet pretrained weights as the starting point for training and trained the model using our collected training data. The final fully connected layer was modified to have 5 outputs for the 5 gaze zones. We trained separate models for each of the four camera views: dashboard-center, dashboard-driver, steering, and rearview. Table 1 shows the classification accuracies with each camera view. Figure 5 shows the confusion matrices for gaze zone classification with the 4 camera views. From the confusion matrices we note that the dashboard driver and steering camera views achieve high accuracies for all gaze zones, compared to the rearview and dashboard-center camera views. Further, the most commonly confused gaze zones are (i) speedometer and forward and (ii) rearview mirror and infotainment. Both of these views correspond to slight differences in head/gaze pitch. The steering column camera which is directly pointed at the driver's face (see Fig. 4) is best suited for distinguishing between these gaze zones.

*Table 1*
*Gaze Zone Classification Accuracies (%)*

| View | EfficientNet-B3 5-class |
| --- | --- |
| Dashboard-driver | 89.71 |
| Dashboard-center | 85.31 |
| Rearview | 83.09 |
| Steering | **91.41** |

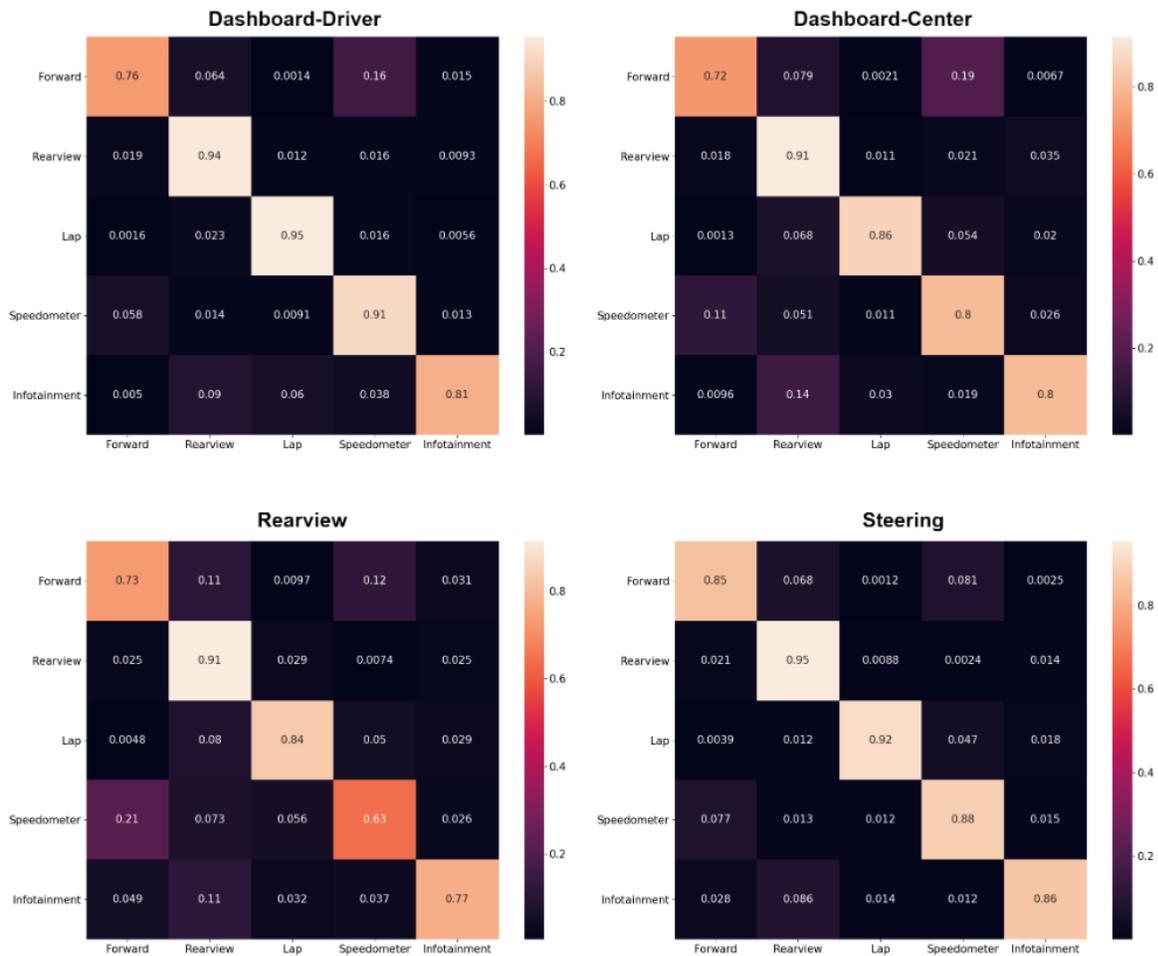

**Figure 5. Confusion matrices for gaze zone classification using the 4 different camera views.**

**Hand Location and Held Object Classification** We use cropped images around the driver's hands to classify the driver's hand locations and held objects using a ResNet-18 model (pretrained on ImageNet). We use a separate model per hand and per function, yielding a four model suite per camera view. Additionally, for each camera view, we use three different crop sizes, as different views may benefit from larger or more constrained contextual regions, and pixel area occupied by the hand may change between views. We sampled from dimensions 50x50, 100x100, and 200x200, and recommend that these hyperparameters can be tuned specific to the camera view and desired task.

With the aforementioned group of combinations, we train a total yield of 48 models. Of these, we select the best-performing model across the crop sizes, with the results reported in the confusion matrices shown in Figures 6 and 7. From these matrices, the dashboard center view provides the best estimate of left and right hand positions, while the rearview and dashboard center cameras appear to perform better for held object classification. For the driver-center view, left hand zone classification accuracy is 79% and right hand zone classification accuracy is 90% corresponding to the confusion matrices in Figure 6. In the rearview and dashboard center cameras respectively, left held object classification accuracy is 75% and right held object classification accuracy is 72% corresponding to the confusion matrices in Figure 7. One promising feature of the held-object classifiers is that the models provide a fairly consistent binary function between whether an object is held or not held (that is, it tends to correctly return "None" when no object is held, and some object when any object is held).

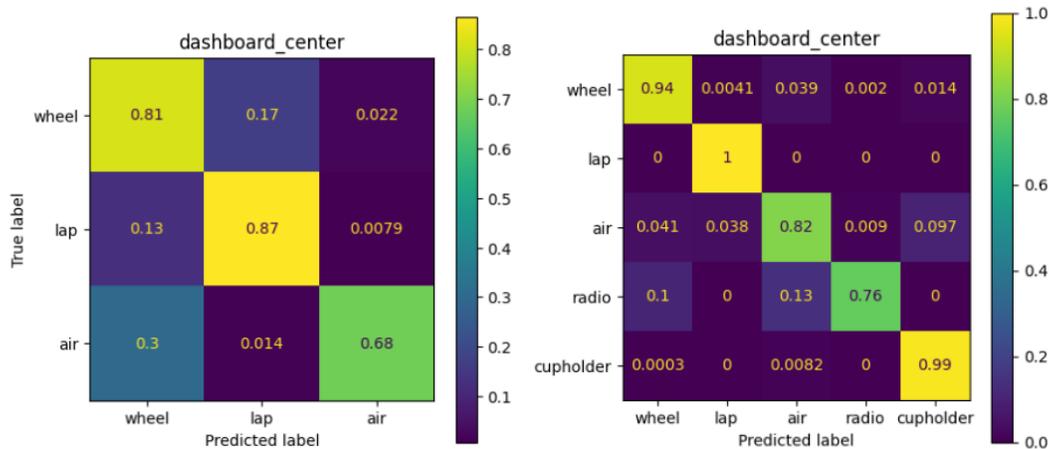

*Figure 6. Confusion matrices for left hand zone classification (left) and right hand zone classification (right) from the experimentally-optimal dashboard center view. Note that the left hand is excluded from reaching the radio or cupholder in our current scheme.*

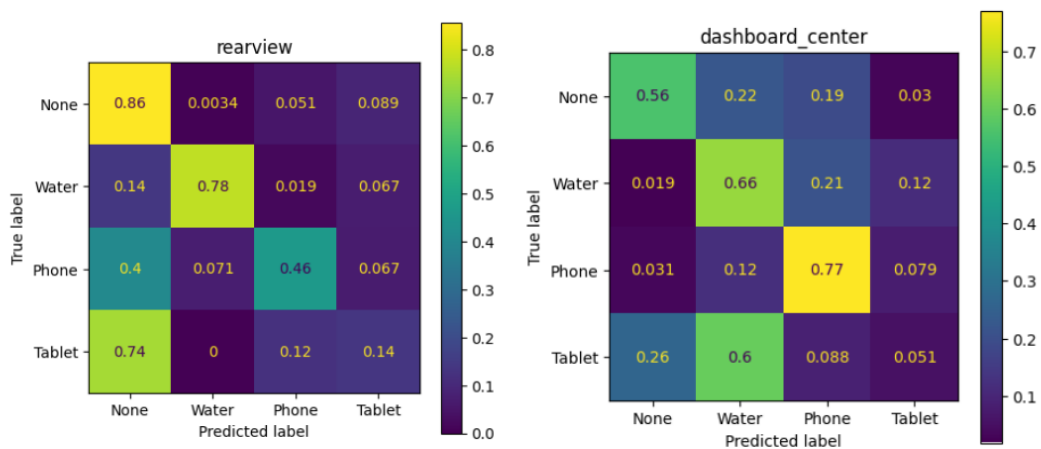

*Figure 7. Confusion matrices for left hand held object classification (left) and right hand object classification (right) from the experimentally optimal camera views.*

**EXPERIMENTS AND EVALUATION**

**Observable Readiness Index Estimation with Multi-Camera Framework for In-Cabin Activity Monitoring**

The Observable Readiness Index (ORI) estimation model consists of two steps, shown in the red blocks of Figure 8. The first step involves extracting frame-wise features capturing the driver's state, and the second step involves using an LSTM model to aggregate temporal context over the past 2 seconds of frame-wise features. The original ORI model described in [15] used frame-wise features from 4 cameras observing the driver's face, hands, feet and body pose, and two IR range sensors observing the driver's hands and feet. To adapt it to the above multi-camera framework, we train the ORI model using a subset of features focused on driver gaze and hand activity, obtained from models described in the above sections for gaze features (gaze zone probabilities) and hand features (held object class probabilities and hand activity class probabilities).

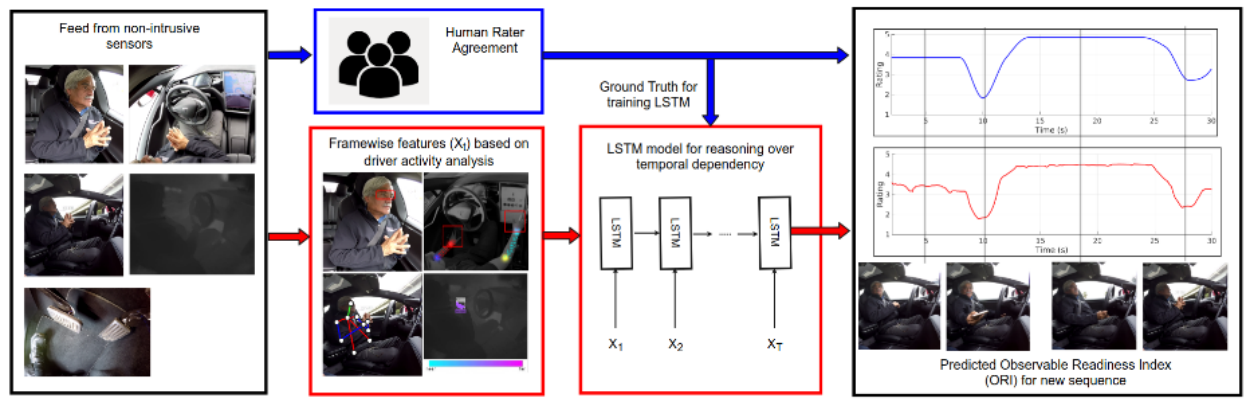

*Figure 8. Model flow for ORI Estimation.*

**Datasets**

Naturalistic driving data for training vision models was collected from two source testbeds: (1) the LISA-T testbed [17] operated in urban and freeway environments in La Jolla, California (USA), and (2) the controlled driving datasets collected from a twin Tesla Model S described in [15, 22, 23], capturing real takeover events from over 100 subjects on a test track in Iowa. Both testbeds have self-driving features which can be operated in real-world freeway environments. Subjects in Iowa performed each of eight different non-driving related tasks (NDRTs) while the Tesla operated on autopilot, maintaining its lane and a cruising speed of 30 mph. Subjects were then issued a TOR while the autopilot was simultaneously disengaged. The drivers are expected to assume control and stabilize the vehicle. Data from the Iowa test track further included distances to lane markings, captured from an onboard Mobileye system, and speed data annotated from the dashboard speedometer.

The LISA-T testbed dataset, used here to evaluate the effectiveness of the multi-camera framework and relationship of readiness to takeover quality, features collected video data on three drivers from four infrared cameras, mounted behind the steering wheel, on the dashboard facing the driver, on the dashboard facing the center of the cabin, and from the rearview mirror facing the center of the cabin (the same positions used in training the gaze and hand models described in the previous section). This naturalistic driving dataset consists of roughly 10 hours of driving data with LISA-T operating in autonomous mode on freeways, with three different drivers. The drivers perform non-driving related tasks (NDRTs) such as operating the infotainment unit to navigate to a specific location, changing the radio station, using a hand-held device to read a text and drinking from cup/bottle. A safety co-passenger constantly monitors the road and vehicle state when the driver performs the NDRT. To simulate takeover requests, the safety co-passenger triggers a TOR, after which the driver is instructed to bring their eyes on the road and hands on the wheel. Note that control is not transferred from the vehicle to the driver after takeover alarms (as in the controlled driving dataset on the Iowa test track) since this would be unsafe in real traffic. Our dataset contains 295 total "takeovers". We extract 30 second clips around each "takeover" event. Two second snippets from these clips are rated by raters on a scale of 1 to 5, as illustrated in Figure 9. The ratings and normalized and interpolated over the video clips to obtain the ground truth ORI ratings.

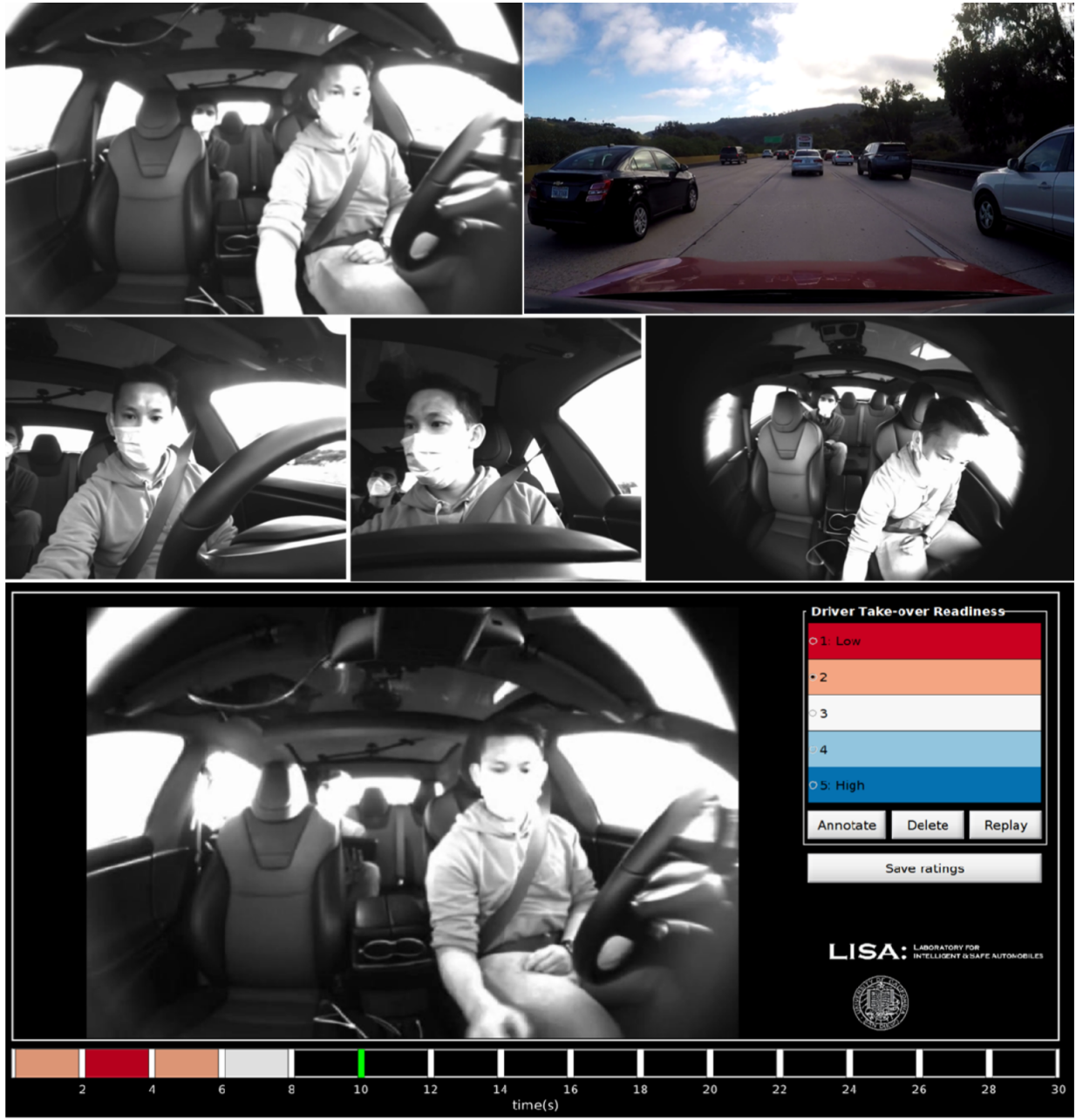

*Figure 9. Raters provide a rating on a scale of 1 to 5 to describe the readiness of the driver over each two second interval of a 30 second driving clip, as described in [1].*

**Observable Readiness Index Estimation**

We train the ORI estimation model using a subset of features (gaze zones, held-objects and hand activity) using naturalistic data from LISA-T and the controlled driving datasets. Note that the extracted features, namely, gaze zone probabilities, held-object probabilities and hand activity probabilities, are all view independent, allowing for training on data from testbeds with differing camera configurations.

We compare three different variants of the ORI model, which use purely gaze features, purely hand features, and both sets of features. Additionally, we compare the camera views used for extracting gaze features and hand features. Table 2 reports the mean absolute errors between the ground truth ORI ratings and the predicted ORI values for the newly collected dataset with the multi-camera framework.

*Table 2*
***Mean average error for ORI estimation under different combinations of features and views***

| Gaze Features | | | | Hand features | | MAE |
|---|---|---|---|---|---|---|
| Dashboard Driver | Dashboard Center | Rearview | Steering | Dashboard Center | Rearview | |
| ✓ | | | | | | **0.6711** |
| | ✓ | | | | | 0.7670 |
| | | ✓ | | | | 0.7404 |
| | | | ✓ | | | 0.6883 |
| | | | | ✓ | | 1.5171 |
| | | | | | ✓ | 1.5722 |
| | ✓ | | | ✓ | | 0.9215 |
| | | ✓ | | | ✓ | 0.9280 |

Experimental results suggest the ORI models that purely use gaze features outperform those that purely use hand features, as well as those that use a combination of both sets of features. In other words, the hand features seem to be adversely affecting ORI estimation. However, previous ORI experiments [15] clearly show the utility of hand locations and held objects for estimating ORI. The poor results with hand features suggest two possibilities: first, that the hand location and held object classifier modules may need to be further trained to improve their accuracies, and second, that the particular selected camera views may not be the most suitable for driver hand analysis, because the driver's hands are only visible in the dashboard-center and rearview cameras (in which views the driver's hands are often occluded or truncated). Within the purely gaze-based ORI estimation models, the models that use the dashboard-mounted driver-facing camera and steering wheel view achieve the lowest MAE values, as expected from views which have the most direct visibility of the eyes.

**Objective Takeover Readiness Metrics Using Ego-Vehicle State**
While the ORI represents prediction of a quantity derived from subjectivity, here we introduce an objective measure of takeover quality. A safe takeover performed by a prepared driver would appear seamless, meaning that the ego vehicle would move in a predictable manner, without sudden braking, acceleration or lateral deviation. A distracted driver, on the other hand, may overreact, leading to unpredictable longitudinal or lateral motion of the ego vehicle. The ego-vehicle's motion during takeovers can thus provide a useful objective measure of driver takeover readiness.

We derive two objective readiness measures from the ego-vehicle state: one corresponding to longitudinal motion, and one corresponding to lateral motion. First, we consider the maximum change in speed ($\Delta v$) of the vehicle during the immediate 5 seconds post TOR, intended to capture any sudden braking or acceleration by an under-prepared driver. Second, we consider maximum deviation from the lane centerline ($\Delta x$) during the immediate 5 seconds post TOR, intended to capture any sudden swerving or drift during the takeover.

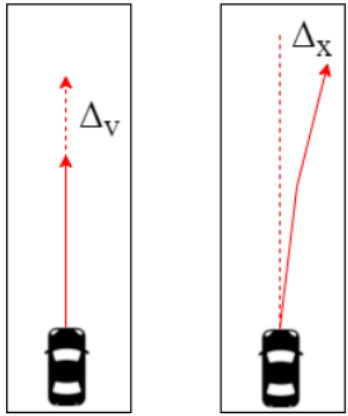

*Figure 10. As objective metrics based on ego-vehicle state, we measure the maximum deviation in ego vehicle speed (Δv) 5 seconds immediately after the TOR (left), and maximum lateral deviation from lane centerline (Δx) 5 seconds immediately after the TOR (right).*

From the collected data, we measure the maximum change in speed, and deviation from lane centerline 5 seconds post TOR for each experimental trial (Δv and Δx). Figures 11 and 12 show the average values of Δv and Δx by NDRT respectively.

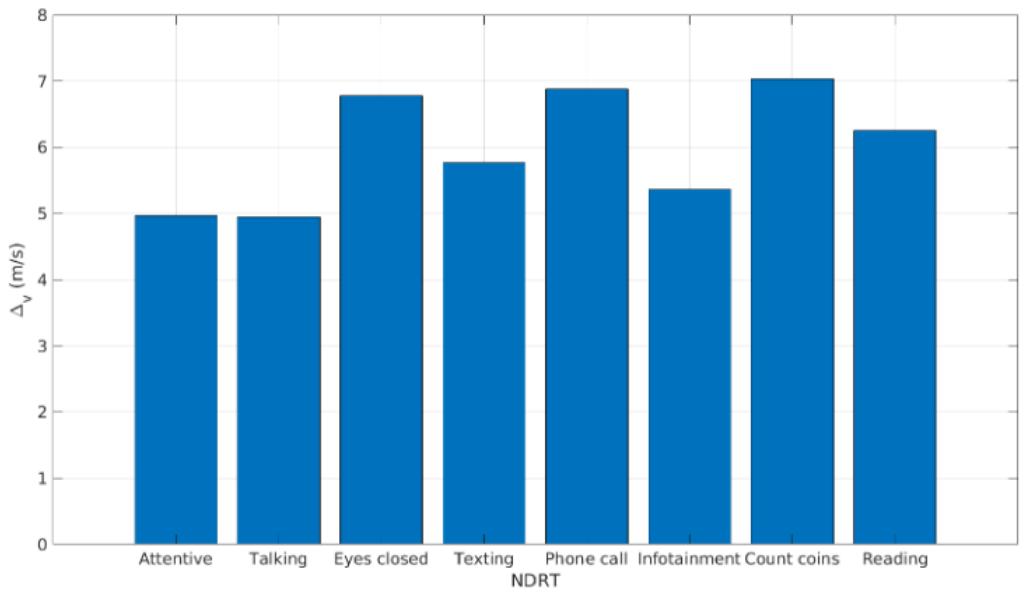

*Figure 11. Max deviation in speed up to 5 seconds post TOR (Δv) by NDRT.*

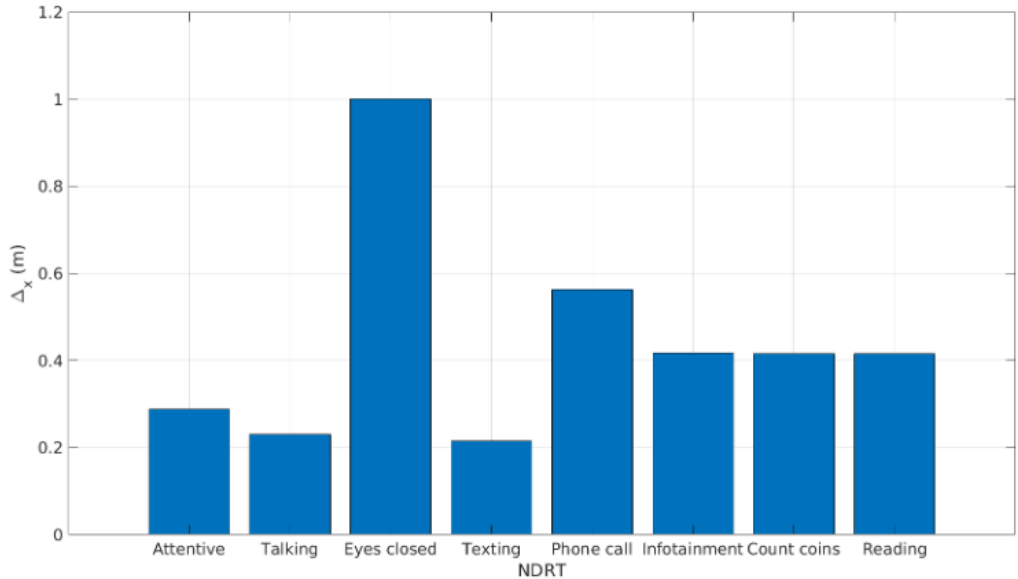

***Figure 12. Max deviation from lane centerline up to 5 seconds post TOR (Δx) by NDRT.***

We note that Δv is lowest when the driver is attentive or just talking to co-passenger, while Δv is high for highly distracting NDRTs such as eyes closed, phone call, reading, counting coins and texting. This suggests that the driver tends to overreact post takeover when distracted. A similar trend can be observed for Δx, with the maximum lateral deviation observed for the eyes closed (sleeping) NDRT.

**Correlation of Objective Readiness Metrics with ORI and Takeover Time**

We compute the correlation of ORI prior to TOR with the objective metrics Δv and Δx, as well as the correlation of Δv and Δx with takeover time, reported in Table 3.

***Table 3***
***Correlation of Δx and Δx with ORI and Takeover Time***

|  | $\Delta_v$ | $\Delta_x$ |
|---|---|---|
| ORI | -0.0467 | -0.0427 |
| Takeover time | 0.1686 | 0.0072 |

We note that we get very slight negative correlations for ORI with Δv and Δx, meaning a low value of observable readiness corresponds to high deviations in speed and high deviations from the lane centerline. On the other hand, takeover times have a slight positive correlation with Δv, meaning high takeover times correspond to high values of Δv . From our experiments, Δx is uncorrelated with takeover time.

**CONCLUDING REMARKS**

This exploration suggests that the ORI model generalizes to various views of the cabin and serves as an effective predictor of TOT, but relies on the performance of the modules for feature extraction from the hands and eyes. Further investigation in methods for improved accuracy of hand zone and held object classifiers would serve to further improve downstream models' (such as ORI) predictive abilities. These experiments suggest that there is a tradeoff in the ability of a camera to accurately observe both the eyes and the hands, and that careful consideration should be made when selecting camera positions intended to predict driver readiness, and that the optimal placement is highly task-dependent.

In analyzing takeover quality as a function of readiness and estimated takeover time, we find the trends in correlation to match intuition; the negative correlations between ORI and takeover quality metrics show that low readiness corresponds to high deviations in vehicle lateral position and longitudinal velocity, and positive correlations between TOT and takeover quality metrics show that quick takeovers correspond to less deviations in vehicle lateral position and longitudinal velocity.

We note the low absolute values of all correlation coefficients, suggesting that the link between the objective readiness metrics and ORI/takeover time is inconclusive. One prominent source of noise is that some drivers brake and bring the ego-vehicle to a halt post TOR in the controlled driving dataset, rather than maintaining ego-vehicle speed. A more precise experiment where the driver is instructed to maintain the ego-vehicle's motion after the TOR may yield more reliable metrics of takeover quality. Additionally, surrounding traffic, which is missing in the controlled driving environment, may affect these values, as would the next navigational goal of the ego vehicle. An initial study varying these parameters in a simulator setting followed by a real world study with an appropriate safety protocol (e.g. secondary driver with pedal controls) could yield results of further interest.

In conclusion, the experimental analysis represents a complete and fully-AI-driven instance of the before-during-after analytical framework for safer control transitions in real-world takeover events; predicted ORI speaks to readiness "before" a takeover, predicted TOT speaks to pace of transition "during" the takeover, and ego vehicle motion metrics speak to the quality of control "after" the takeover.

**REFERENCES**

[1] Møgelmose, A., Trivedi, M. M., & Moeslund, T. B. (2015, June). Trajectory analysis and prediction for improved pedestrian safety: Integrated framework and evaluations. In 2015 IEEE intelligent vehicles symposium (IV) (pp. 330-335). IEEE.

[2] Greer, R., & Trivedi, M. (2022). From Pedestrian Detection to Crosswalk Estimation: An EM Algorithm and Analysis on Diverse Datasets. arXiv preprint arXiv:2205.12579.

[3] Trivedi, M. M., Gandhi, T. L., & Huang, K. S. (2005). Distributed interactive video arrays for event capture and enhanced situational awareness. IEEE Intelligent Systems, 20(5), 58-66.

[4] Greer, R., Deo, N., & Trivedi, M. (2021). Trajectory prediction in autonomous driving with a lane heading auxiliary loss. IEEE Robotics and Automation Letters, 6(3), 4907-4914.

[5] Ohn-Bar, E., Tawari, A., Martin, S., & Trivedi, M. M. (2015). On surveillance for safety critical events: In-vehicle video networks for predictive driver assistance systems. Computer Vision and Image Understanding, 134, 130-140.

[6] Cheng, S. Y., Park, S., & Trivedi, M. M. (2007). Multiperspective and multimodal video arrays for 3D body tracking and activity analysis. Comput. Vis. Image Underst.(Special Issue on Advances in Vision Algorithms and Systems Beyond the Visible Spectrum), 106(2-3), 245-257.

[7] Martin, S., Tawari, A., & Trivedi, M. M. (2014). Toward privacy-protecting safety systems for naturalistic driving videos. IEEE Transactions on Intelligent Transportation Systems, 15(4), 1811-1822.

[8] Cummings, M. L., & Bauchwitz, B. (2021). Safety Implications of Variability in Autonomous Driving Assist Alerting. IEEE Transactions on Intelligent Transportation Systems.

[9] Karakaya, B., & Bengler, K. (2021, June). Investigation of driver behavior during minimal risk maneuvers of automated vehicles. In Congress of the International Ergonomics Association (pp. 691-700). Springer, Cham.

[10] Naujoks, F., Hergeth, S., Keinath, A., Wiedemann, K., & Schömig, N. (2019). Development and Application of an Expert Assessment Method for Evaluating the Usability of SAE Level 3 Ads HMIs. System, 3, L2.

[11] Daman, P., Götze, M., Gold, C., & Kompass, K. (2019). BMW's Safety Guidelines for the Testing and Deployment of Automated Vehicles. In 26th International Technical Conference on the Enhanced Safety of Vehicles (ESV): Technology: Enabling a Safer TomorrowNational Highway Traffic Safety Administration (No. 19-0226).

[12] Kurpiers, C., Lechner, D., & Raisch, F. (2019, June). The influence of a gaze direction based attention request to maintain mode awareness. In Proceedings of the 26th International Technical Conference on the Enhanced Safety of Vehicles, Eindhoven, The Netherlands (pp. 10-13).

[13] National Highway Traffic Safety Administration. (2017). Automated driving systems 2.0: A vision for safety. Washington, DC: US Department of Transportation, DOT HS, 812, 442.

[14] Thorn, E., Kimmel, S. C., Chaka, M., & Hamilton, B. A. (2018). A framework for automated driving system testable cases and scenarios (No. DOT HS 812 623). United States. Department of Transportation. National Highway Traffic Safety Administration.

[15] Deo, N., Trivedi, M.M.: Looking at the driver/rider in autonomous vehicles to predict take-over readiness. IEEE Transactions on Intelligent Vehicles 5(1), 41–52 (2019)

[16] Victor, T., Dozza, M., Bärgman, J., Boda, C. N., Engström, J., Flannagan, C., ... & Markkula, G. (2015). Analysis of naturalistic driving study data: Safer glances, driver inattention, and crash risk (No. SHRP 2 Report S2-S08A-RW-1).

[17] Rangesh, A., Deo, N., Yuen, K., Pirozhenko, K., Gunaratne, P., Toyoda, H., & Trivedi, M. M. (2018, November). Exploring the situational awareness of humans inside autonomous vehicles. In 2018 21st International Conference on Intelligent Transportation Systems (ITSC) (pp. 190-197). IEEE.

[18] Rangesh, A., Trivedi, M.M.: Handynet: A one-stop solution to detect, segment, localize & analyze driver hands. In: Proceedings of the IEEE Conference on Computer Vision and Pattern Recognition Workshops. pp. 1103–1110 (2018)

[19] Yuen, K., Trivedi, M.M.: Looking at hands in autonomous vehicles: A convnet approach using part affinity fields. IEEE Transactions on Intelligent Vehicles 5(3), 361–371 (2019)

[20] Vora, S., Rangesh, A., and Trivedi, M.M.: "Driver Gaze Zone Estimation using Convolutional Neural Networks: A General Framework and Ablative Analysis," IEEE Transactions on Intelligent Vehicles, 2018.

[21] Rangesh, A., and Trivedi, M.M.: "Forced Spatial Attention for Driver Foot Activity Classification," ICCV Workshop on Assistive Computer Vision and Robotics (Oral), 2019.

[22] Rangesh, A., Deo, N., Greer, R., Gunaratne, P., Trivedi, M.M.: Autonomous vehicles that alert humans to take-over controls: Modeling with real-world data. In: 2021 IEEE International Intelligent Transportation Systems Conference (ITSC). pp. 231–236. IEEE (2021)

[23] Rangesh, A., Deo, N., Greer, R., Gunaratne, P., Trivedi, M.M.: Predicting take-over time for autonomous driving with real-world data: Robust data augmentation, models, and evaluation. arXiv preprint arXiv:2107.12932 (2021)

[24] Shi, E., & Bengler, K. (2022). Non-driving related tasks' effects on takeover and manual driving behavior in a real driving setting: A differentiation approach based on task switching and modality shifting. Accident Analysis & Prevention, 178, 106844.

[25] Zeeb, K., Buchner, A., & Schrauf, M. (2016). Is take-over time all that matters? The impact of visual-cognitive load on driver take-over quality after conditionally automated driving. Accident analysis & prevention, 92, 230-239.

[26] Chen, K., Wang, J., Pang, J., Cao, Y., Xiong, Y., Li, X., Sun, S., Feng, W., Liu, Z., Xu, J., Zhang, Z., Cheng, D., Zhu, C., Cheng, T., Zhao, Q., Li, B., Lu, X., Zhu, R., Wu, Y., Dai, J., Wang, J., Shi, J., Ouyang, W., Loy, C., and Lin, D.: Open MMLab Detection Toolbox and Benchmark. arXiv preprint arXiv:1906.07155 (2019)

[27] Ren, S., He, K., Girshick, R., & Sun, J. (2015). Faster r-cnn: Towards real-time object detection with region proposal networks. Advances in neural information processing systems, 28.

[28] Lin, T. Y., Dollár, P., Girshick, R., He, K., Hariharan, B., & Belongie, S. (2017). Feature pyramid networks for object detection. In Proceedings of the IEEE conference on computer vision and pattern recognition (pp. 2117-2125).

[29] He, K., Zhang, X., Ren, S., & Sun, J. (2016). Deep residual learning for image recognition. In Proceedings of the IEEE conference on computer vision and pattern recognition (pp. 770-778).

[30] Lin, T. Y., Maire, M., Belongie, S., Hays, J., Perona, P., Ramanan, D., ... & Zitnick, C. L. (2014, September). Microsoft coco: Common objects in context. In European conference on computer vision (pp. 740-755). Springer, Cham.

[31] MMPose Contributors: OpenMMLab Pose Estimation Toolbox and Benchmark, https://github.com/open-mmlab/mmpose (2020)

[32] Sun, K., Xiao, B., Liu, D., & Wang, J. (2019). Deep high-resolution representation learning for human pose estimation. In Proceedings of the IEEE/CVF conference on computer vision and pattern recognition (pp. 5693-5703).

[33] Tan, M., & Le, Q. (2019, May). Efficientnet: Rethinking model scaling for convolutional neural networks. In International conference on machine learning (pp. 6105-6114). PMLR.